\pdfoutput=1

\documentclass[11pt, reqno, dvipsnames]{article}

\usepackage{emnlp2021}

\usepackage{times}
\usepackage{latexsym}

\usepackage{float}
\usepackage{booktabs}
\usepackage{xcolor}
\usepackage{graphicx}
\usepackage{gensymb} 
\usepackage{subfig}
\usepackage{caption}
\usepackage{ragged2e}

\usepackage{todonotes}
\makeatletter
\newcommand*\iftodonotes{\if@todonotes@disabled\expandafter\@secondoftwo\else\expandafter\@firstoftwo\fi}  
\makeatother

\usepackage[T1]{fontenc}

\usepackage[utf8]{inputenc}

\usepackage{microtype}

\usepackage{scalerel, stackengine, amsmath}
\newcommand\equalhat{\mathrel{\stackon[1.5pt]{=}{\stretchto{%
    \scalerel*[\widthof{=}]{\wedge}{\rule{1ex}{3ex}}}{0.5ex}}}}

\usepackage{amsfonts}
\usepackage{mathtools}
\DeclarePairedDelimiter\floor{\lfloor}{\rfloor}

\usepackage{enumitem}

\title{A Massively Multilingual Analysis of Cross-linguality\\ in Shared Embedding Space}

\author{Alex Jones \\ Dartmouth College \\ \href{mailto:alexander.g.jones.23@dartmouth.edu}{\tt{\color{black}{alexander.g.jones.23@dartmouth.edu}}}
        \AND
        William Yang Wang \\ University of California, Santa Barbara \\ \href{mailto:william@cs.ucsb.edu}{\tt{\color{black}{william@cs.ucsb.edu}}}
        \And
        Kyle Mahowald \\ University of Texas at Austin \\ \href{mailto:mahowald@ucsb.edu}{\tt{\color{black}{mahowald@utexas.edu}}}}

\begin{document}
\maketitle
\begin{abstract}
In cross-lingual language models, representations for many different languages live in the same space. 
Here, we investigate the linguistic and non-linguistic factors affecting sentence-level alignment in cross-lingual pretrained language models for 101 languages and 5,050 language pairs. Using BERT-based LaBSE and BiLSTM-based LASER as our models, and the Bible as our corpus, we compute a task-based measure of cross-lingual alignment in the form of bitext retrieval performance, as well as four intrinsic measures of vector space alignment and isomorphism. We then examine a range of linguistic, quasi-linguistic, and training-related features as potential predictors of these alignment metrics. The results of our analyses show that \textit{word order agreement} and \textit{agreement in morphological complexity} are two of the strongest linguistic predictors of cross-linguality. We also note \textit{in-family} training data as a stronger predictor than \textit{language-specific} training data across the board. We verify some of our linguistic findings by looking at the effect of morphological segmentation on English-Inuktitut alignment, in addition to examining the effect of word order agreement on isomorphism for 66 zero-shot language pairs from a different corpus. We make the data and code for our experiments publicly available.\footnote{\url{https://github.com/AlexJonesNLP/XLAnalysis5K}}

\end{abstract}

\section{Introduction}

Cross-lingual language models are polyglots insofar as they house representations for many different languages in the same space. But to what extent are they \textit{good} polyglots? The answer depends, in part, on how well-aligned and isomorphic the representations are, and not all language pairs are equally well-aligned. What determines the quality of the alignment? Are language pairs from the same family (e.g., Spanish and French) better aligned than languages from two unrelated families (e.g., Japanese and Swahili)? Are languages which are geographically closer or share an alphabet better aligned? How do factors from linguistic typology (like word order and morphological marking) affect alignment?

Recent work has looked at the typological and training-related factors affecting cross-lingual alignment in monolingual embedding space \cite{vulic-etal-2020-good, dubossarsky-etal-2020-secret}, assessed the cross-linguality of pretrained language models using probing tasks and downstream performance measures \cite{conneau-etal-2020-emerging, wu-dredze-2019-beto, wu-dredze-2020-languages, pires-etal-2019-multilingual, groenwold-etal-2020-evaluating}, and probed Transformer models \cite{wolf-etal-2020-transformers} for linguistic structure (see \citealt{rogers-etal-2020-primer} for an overview of  over 150 studies). However, a gap in the research exists regarding the following question: What are the linguistic, quasi-linguistic, and training-related factors determining the cross-linguality of \textit{sentence representations} in \textit{shared embedding space}, and what are the relative weights of these factors?

\begin{figure}
    \centering
    \includegraphics[scale=0.35]{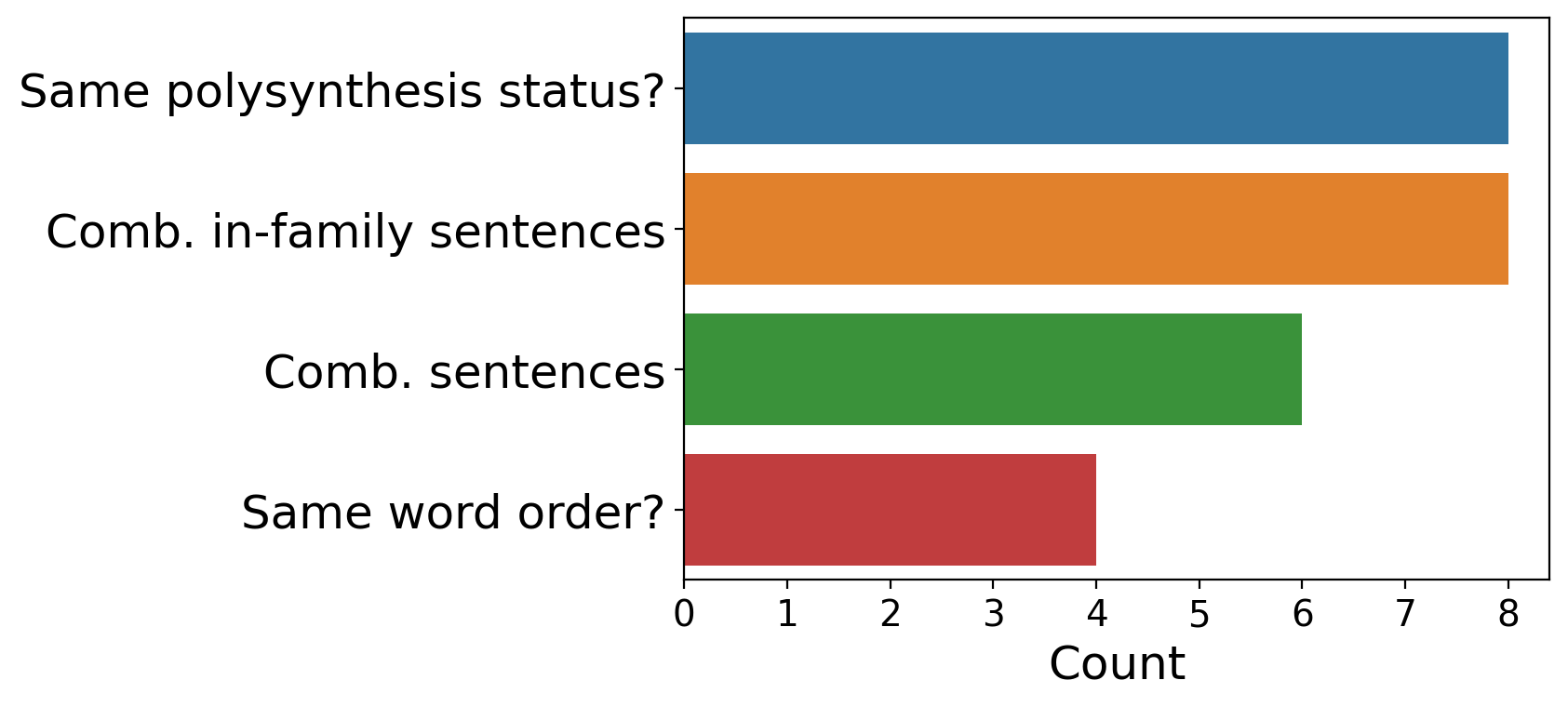}
    \caption{A look at some of the strongest features for predicting cross-linguality, according to their number of occurrences in best-feature regression searches across all our dependent variables (see Section \ref{feature_search}).}
    \label{fig:sneak_peak}
\end{figure}


We argue, that given the importance of alignment in multilingual model performance, gaining fundamental insight into what affects inter-language alignment and isomorphism---specifically by exploring which \textbf{linguistic} factors matter for inter-language alignment---will make it possible to leverage existing information on linguistic typology to improve alignment (and thereby task performance) for low-resource languages. 

Our contributions are as follows:
\begin{itemize}
    \item We provide a characterization of cross-linguality for 101 languages (29 language families) in two massively multilingual sentence embedding models with different architectures (LaBSE and LASER), attacking the question from the vantage of vector space analysis---using four measures of alignment and isomorphism---and downstream task performance (namely bitext retrieval).
    \item We present over a dozen linguistic, quasi-linguistic, and training-related factors as potential predictors of cross-linguality, and examine their relationship with the above metrics using diverse statistical analyses.
    \item We uncover novel and pronounced effects of \textit{morphology agreement} and \textit{word order agreement} on cross-linguality, demonstrate the importance of \textit{in-family training data} in ensuring multilinguality, and validate our linguistic findings with two empirical case studies on  low-resource languages.
\end{itemize}

\section{Related Work}

Various studies have assessed the cross-linguality of pretrained language models. Recent efforts have approached this question via performance on an array of downstream NLP tasks \cite{conneau-etal-2020-emerging, wu-dredze-2019-beto, wu-dredze-2020-languages, karthikeyan-etal-2020-cross, pires-etal-2019-multilingual, groenwold-etal-2020-evaluating}, and others have proposed methods for better cross-lingual alignment in light of systematic cross-lingual deficiencies \cite{zhang-etal-2019-girls, xia-etal-2021-metaxl}. Our study hews closest methodologically to \citet{vulic-etal-2020-good} and \citet{dubossarsky-etal-2020-secret}, who investigate the determinants of cross-lingual isomorphism using monolingual \texttt{fastText} embeddings \cite{bojanowski-etal-2016-enriching, joulin-etal-2016-bag, mikolov-etal-2013-efficient}.

Findings from these studies have been mixed, but some patterns emerge. \citet{pires-etal-2019-multilingual} and \citet{conneau-etal-2020-emerging} find that cross-lingual transfer works best between typologically similar language pairs, in particular between languages that share word order features. \citet{wu-dredze-2019-beto} approach cross-linguality by focusing on zero-shot cross-lingual transfer in mBERT, and show that each mBERT layer retains language-specific information and that token overlap correlates with cross-lingual performance. \citet{wu-dredze-2020-languages} home in on low-resource languages, finding that they often fail to reap the benefits of massively multilingual joint training but that their performance can be boosted by providing similar-language training data. Somewhat contrary to others' results (including ours), \citet{karthikeyan-etal-2020-cross} find that lexical overlap factors in negligibly to cross-lingual transfer, while the depth of the network is integrally important. \citet{vulic-etal-2020-good} and \citet{dubossarsky-etal-2020-secret} look at how typological features, training-related factors, and measures of vector space isomorphism predict cross-lingual performance between monolingual word embeddings. \citet{vulic-etal-2020-good} find in their experiments that cross-lingual performance depends mostly on training data and regimes, while \citet{dubossarsky-etal-2020-secret} see more mixed results from their experiments: They show that linguistic typology is important, but not deterministic, for predicting cross-lingual performance. 

Our work not only replicates these findings for monolingual spaces in the multilingual embedding space (e.g. on word order similarity, related training data, typological distance, subword overlap), but extends that work through: (1) The \textbf{scale} (101 languages, 5,050 language pairs in the main analysis); (2) The \textbf{quantity and diversity of predictors} (13 linguistic, quasi-linguistic, and training-related features); (3) The models (cross-lingual \textbf{sentence} encoders with \textbf{different architectures}); and (4) The analytic methods (a blend of \textbf{prediction-based} and \textbf{classical statistical} techniques, supplemented by performance-based \textbf{case studies} on extremely low-resource languages).



\section{Bible Corpus}

The source of the bitexts we evaluate on is the superparallel Bible corpus\footnote{\url{http://christos-c.com/bible/}. We use this corpus because it is massively multilingual and well-aligned.} from \citet{christodouloupoulos_steedman_2014}, whence we gather texts for 101 languages and bitexts for 5,050 language pairs.\footnote{${101\choose2}=5050$} We evaluate on the Books of Matthew and John in the New Testament separately and average the results, as these parts are available for all 101 languages. In doing so, we avoid the pitfalls of relying on a single set of bitexts for our analysis. Each document contains 800-1000 sentences.

\section{Measures of Cross-lingual Alignment \& Isomorphism}

We formulate alignment metrics in two distinct ways: over \textit{language pairs} and over \textit{individual languages}. The latter group is computed from the former by averaging over all pairs in which a language appears. For example, to derive the average F1-score for Chinese, we average over the F1-scores for Chinese-German, Chinese-Amuzgo, etc.

Some metrics we use are measures of vector subspace \textit{isomorphism} (i.e. those examined in \citealt{dubossarsky-etal-2020-secret}), while others are measures of \textit{alignment} (namely those pertaining to bitext retrieval). Vector spaces may be isomorphic without being well-aligned, so we quantify multilinguality in diverse ways.

\subsection{Bitext Retrieval Task}\label{bitext-retrieval}

The bitext retrieval task consists of finding all sentences in a paired set of documents that are translations of each other. This process can be carried out between two \textit{comparable} corpora, such as Wikipedia (``bitext mining''), but we use the 5,050 bitexts collected from the Bible corpus. We mine in two directions: for each sentence in document $\mathcal{X}$, we find a match in document $\mathcal{Y}$, and vice-versa. We then take the intersection of those two searches, which has proven to be a useful heuristic \cite{artetxe-schwenk-2019-margin, jones-wijaya-2021-majority}.
Note that this task can be thought of as the sentence-level analog to the bilingual lexicon induction (BLI) task used in \citet{vulic-etal-2020-good} and \citet{dubossarsky-etal-2020-secret}.

\paragraph{Task performance}

Margin scoring, introduced by \citet{artetxe-schwenk-2019-margin}, has shown success on the bitext retrieval task \cite{schwenk-etal-2021-wikimatrix, schwenk-etal-2019-ccmatrix, keung-etal-2021-unsup, tran2020cross, fan-etal-2020-beyond, jones-wijaya-2021-majority}. Margin score may be thought of as ``relativized'' cosine similarity, in that it selects vectors that ``stand out'' most among their neighbors in terms of proximity, rather than ones that are simply closest together. The method requires initially finding the $k$-nearest neighbors of each source and target sentence, which we do efficiently with \texttt{Faiss} \cite{johnson-etal-2017-faiss}. The sentence pair $(x,y)$ is then chosen to maximize the \textit{margin score} between $x$ and $y$, namely
\begin{equation}\label{eq:margin_score}
\begin{split}
&\mathrm{score_{margin}}(x,y) = \\
&\frac{2k\cos{(x,y)}}{\sum_{z\in NN_{k}(x)}{\cos{(x,z)}}+\sum_{z\in NN_{k}(y)}{\cos{(y,z)}}}\nonumber
\end{split} 
\end{equation}

\noindent After retrieving sentence pairs in both directions and keeping the intersection, we compute standard F1-score against ground-truth alignments.

\paragraph{Average margin score}

We also introduce a novel alignment metric in the form of the \textit{average margin score} across ground-truth sentence alignments. Namely, given aligned sentence embedding matrices $\mathcal{X}$ and $\mathcal{Y}$ with $N$ embeddings each, the average margin score is computed as
\begin{equation}\label{eq:avg_margin}
\begin{split}
&\mathrm{margin_{avg}}(\mathcal{X},\mathcal{Y}) = \\ &\frac{1}{N}\sum_{i=1}^{N}{\mathrm{score_{margin}}(\mathcal{X}_{i},\mathcal{Y}_{i}}) \mid \mathcal{X}, \mathcal{Y} \in \mathbb{R}^{N \times emb\_dim}\nonumber
\end{split} 
\end{equation}

\noindent This provides a continuous measure of cross-lingual alignment that is correlated with, but not equivalent to, the F1-score on this task.

\subsection{Approximate Isomorphism}
\citet{vulic-etal-2020-good} and \citet{dubossarsky-etal-2020-secret} introduce various ways of quantifying the degree of isomorphism between two vector spaces, of which we use three. Note that unlike \citet{vulic-etal-2020-good} and \citet{dubossarsky-etal-2020-secret}, who investigate isomorphism between \textit{monolingual} spaces, we examine cross-lingual isomorphism within \textit{shared} embedding space. These metrics thus technically quantify vector \textit{subspace} isomorphism, where each subspace comprises embeddings in a particular language.

\paragraph{Gromov-Hausdorff distance}

The Hausdorff distance between two metric spaces $\mathcal{X}$ and $\mathcal{Y}$, given by

\begin{align*}
\mathcal{H}(\mathcal{X},\mathcal{Y}) = \max[{\sup_{x \in \mathcal{X}}{\inf_{y \in \mathcal{Y}}}{d(x,y), \sup_{y \in \mathcal{Y}}{\inf_{x \in \mathcal{X}}}{d(x,y)}}}]
\end{align*}

\noindent intuitively measures the worst-case distance between the nearest neighbors of $\mathcal{X}$ and $\mathcal{Y}$ \cite{vulic-etal-2020-good}. The \textit{Gromov-Hausdorff distance} then minimizes this distance over all isometric transforms $f$ and $g$:

\begin{align*}
\mathcal{GH}(\mathcal{X},\mathcal{Y}) = \inf_{f,g}{\mathcal{H}(f(\mathcal{X}),g(\mathcal{Y}))}
\end{align*}

\noindent In practice, the Gromov-Hausdorff distance is approximated by computing the Bottleneck distance between $\mathcal{X}$ and $\mathcal{Y}$ \cite{dubossarsky-etal-2020-secret, chazal-etal-2009-gromov}.

\paragraph{Singular value gap}

Given cross-lingual aligned sentence embeddings stored in matrices $\mathcal{X}$ and $\mathcal{Y}$, each with $n$ singular values $\sigma_{1}, \sigma_{2}, . . . , \sigma_{n}$ sorted in descending order, the \textit{singular value gap} \cite{dubossarsky-etal-2020-secret} between $\mathcal{X}$ and $\mathcal{Y}$ is defined as
\begin{align*}
\mathrm{SVG}(\mathcal{X},\mathcal{Y}) = \sum_{i=1}^{n}{(\log{\sigma_{i}^{\mathcal{X}}} - \log{\sigma_{i}^{\mathcal{Y}})^2}}
\end{align*}

\paragraph{Effective condition number}


The \textit{effective condition number} \cite{dubossarsky-etal-2020-secret} of a matrix $\mathcal{X}$ intuitively captures the extent to which small perturbations in $\mathcal{X}$ are amplified as a result of arbitrary transformations $\phi({\mathcal{X}})$. The lower the (effective) condition number of an embedding space, the more robust it is to transformations (e.g. transfer functions mapping one embedding space to another).

\citet{dubossarsky-etal-2020-secret} reason that monolingual embedding spaces with lower (effective) condition numbers map better to other spaces. They further show that taking the harmonic mean of the effective condition numbers (ECOND-HM) of two embedding spaces provides a reliable measure of approximate isomorphism between those spaces\footnote{As validated by performance on downstream tasks.}. We use ECOND-HM in a similar fashion to gauge the approximate isomorphism, or ``mappability,'' of cross-lingual embedding subspaces, where a lower ECOND-HM indicates greater isomorphism.



\section{Predictors}

\subsection{Linguistic Features}

Similarly to the alignment metrics, we define separate sets of features pertaining to \textit{language pairs} and pertaining to \textit{individual languages}. We take note of this in our descriptions below.

\paragraph{Phylogeny}

For individual languages (all languages in the New Testament corpus), we use both language \textit{family} and \textit{subfamily} as categorical features. For language pairs, we define two binary variables: \textit{same family} and \textit{same subfamily}, corresponding to whether two languages are in the same family or subfamily, respectively.

We include \textit{subfamily} as a feature in order to investigate finer-grained typological and phylogenetic differences that may affect cross-lingual alignment or isomorphism.


\paragraph{Word order typology}
For individual languages, we include \textit{basic word order} as a feature, using the canonical six-way taxonomy (i.e. permutations of \{S, O, V\}). For language pairs, we define binary feature \textit{same word order} analogously to the binary features above. 
We consult the  \texttt{WALS} database\footnote{\url{https://wals.info}} \cite{wals} and Glottolog\footnote{\url{https://glottolog.org}} \cite{hammarstrom-etal-2020-glottolog} to assign dominant word orders.


\paragraph{Morphological typology}

Though it is possible to make fine-grained distinctions in morphological typology in theory, we simply draw a binary distinction between languages that are widely considered \textit{polysynthetic} (mostly Amerindian languages) and all other languages. Even more so than word order, morphological complexity is gradient \citep{cotterell-etal-2019-complexity}. But we argue that polysynthetic languages pose a unique challenge for NLP systems and so perform one-vs-all binary coding such that individual languages are associated with a \textit{polysynthesis status} and language pairs are associated with the feature \textit{same polysynthesis status}. We classify 17 languages in the corpus as polysynthetic.


\paragraph{Typological distance}

We also use typological word vectors from \texttt{lang2vec}\footnote{\url{https://github.com/antonisa/lang2vec}} \cite{malaviya-etal-2017-learning}, based on the \texttt{URIEL}\footnote{\url{http://www.cs.cmu.edu/~dmortens/projects/7_project/}} typological database \cite{littell-etal-2017-uriel} to compute the distance between languages on the basis of aggregated linguistic features. Specifically, we compute:
\begin{enumerate}[nosep]
    \item \textit{Syntactic distance} using KNN-based syntax vectors
    \item \textit{Phonological distance} using KNN-based phonology vectors
    \item \textit{Inventory distance} using KNN-based phonological inventory vectors (distinct from phonological distance)
    \item \textit{Geographic distance} using geographic location vectors
\end{enumerate}
All distances are computed as cosine distances.


\paragraph{Character- \& token-level overlap}
The standard Jaccard similarity coefficient quantifies the overlap between sets $\mathbf{A}$ and $\mathbf{B}$ as:
\begin{align*}
J(\mathbf{A},\mathbf{B}) = \frac{|\mathbf{A}\bigcap\mathbf{B}|}{|\mathbf{A}\bigcup\mathbf{B}|}
\end{align*}
However, this measure fails to take into account the frequency of the items (here, characters) in each set. What we really want is the \textit{weighted}, or \textit{multiset}, version of the Jaccard coefficient. For our purposes, it suffices to reformulate $J$ as:
\begin{equation}\label{eq:weighted_jaccard}
\begin{split}
J_{M}(\mathcal{X},\mathcal{Y})=\frac{|chr(\mathcal{X}_{M}) \bigcap chr(\mathcal{Y}_{M})|}{|chr(\mathcal{X}_{M})\bigcup chr(\mathcal{Y}_{M})|} \forall \mathcal{X}, \mathcal{Y} \in \mathbf{C}\nonumber
\end{split}\raisetag{2\baselineskip}
\end{equation}
where $chr(\mathcal{D}_{M})$ represents the multiset of characters in document $\mathcal{D}$, and $\mathbf{C}$ is the corpus of bitexts we're working with. For convenience and to avoid redundancy, we compute $J_M$ (\textit{character-level overlap}) only on aligned texts in the Book of Matthew.
\textit{Token-level overlap} is computed analogously, using the wordpiece \cite{wu-etal-2016-google} tokenization method employed by LaBSE\footnote{\url{https://huggingface.co/sentence-transformers/LaBSE}}. This measure is only computed on texts in the Book of John.

\subsection{Training-related Features}

The aim of our analysis is to understand the effect of each of the previously described features on cross-lingual alignment and isomorphism \textit{when training factors are controlled for}. To this end, we control for several (pre)training data quantities for the models tested.

First, we account for \textit{language-specific training data} for individual languages. However, we also account for \textit{combined language-specific training data} for language \textit{pairs}, i.e. the amount of data for $x$ plus the amount of data for $y$, where $(x,y)$ is a language pair. We then take it a step further and record \textit{(combined) in-family training data} and \textit{(combined) in-subfamily training data}, taking inspiration from gains made using transfer languages for cross-lingual learning \cite{johnson-etal-2017-googles, littell-etal-2019-multi, lin-etal-2019-choosing}.

By considering these broader training-related statistics, we are able to better control for and observe the role higher-level typological information (e.g. at the family or subfamily level) plays in training these models.



\section{Analysis}

\subsection{Simple Correlations}


\paragraph{Training data}
We first look at simple correlations between the training data quantities and the dependent variables (measures of alignment/isomorphism). Results for language \textit{pairs} are given across all dependent variables for LaBSE and LASER in Table \ref{tab:train_corr}. The most striking observation is that \textit{combined in-family training data} is more highly correlated\footnote{In terms of magnitude; the direction is determined by whether a given metric is measuring cross-linguality positively or negatively.} with the dependent variables than simple \textit{combined data} or \textit{combined in-subfamily data} for \textbf{all} dependent variables, for both LaBSE and LASER\footnote{These results hold even when the effect of combined data on the DV is held constant. Computing the semi-partial correlation \citep{abdi-2007-part} between combined in-family sentences and each DV with combined sentences as the $y$-covariate, we see a change of $r = -0.05$ for LaBSE and $r = +0.004$ for LASER relative to the simple correlations.} ($0.12\leq |r| \leq 0.57$)\footnote{Here, $|\cdot|$ is the absolute value operator.}. At the individual language level, results are similar (i.e. in-family data is most significant), but with weaker correlations across the board ($0.02\leq|r|\leq0.18$). Based on these preliminary results, we highlight \textit{combined in-family training data} as a moderately strong predictor of alignment/isomorphism for a given language pair, one that is in fact \textit{better} than language-specific data for making predictions about massively multilingual sentence models.


\paragraph{(Quasi)-linguistic Features}

Among the predictors, there were several noteworthy correlations. \textit{Same family} was moderately correlated with better alignment/isomorphism in both LaBSE and LASER (generally $0.2<|r|<0.45$), while \textit{same subfamily} was somewhat less correlated. This informs us as to the level at which related-language data is useful for building massively cross-lingual models. \textit{Same word order} and \textit{same polysynthesis status} had comparable relationships with the dependent variables as did \textit{same family}. \textit{Token-level overlap} was moderately but inconsistently correlated with dependent variables ($\approx 0.05 < |r| < 0.5$), while 
\textit{character-level overlap} was somewhat more weakly correlated. The \textit{typological distance} features were weakly but non-negligibly correlated with dependent variables ($\approx 0.1 < |r| < 0.3$), with one outlier (\textit{syntactic distance} was correlated with $r=-0.44$ with bitext retrieval F1-score for LASER). The typological distance features were moderately correlated with one another.



\begin{table*}[h]
    \centering
    \begin{tabular}{llll}
    & \textit{Comb. sentences} & \textit{Comb. in-family} & \textit{Comb. in-subfamily} \\
    \textbf{Metric} && \textit{sentences} & \textit{sentences} \\\toprule
    & LaBSE LASER & LaBSE LASER & LaBSE LASER \\\hline
    Bitext retrieval (F1) & \hspace{1pt} 0.34 \hspace{16pt} 0.13 & \hspace{1pt} \textbf{0.49} \hspace{13pt} \hspace{1pt} \textbf{0.57} & \hspace{1pt} 0.46 \hspace{16pt} 0.35 \\
    Avg. margin score & \hspace{1pt} 0.30 \hspace{13pt} -0.03 & \hspace{1pt} \textbf{0.40} \hspace{16pt} \textbf{0.14} & \hspace{1pt} 0.37 \hspace{16pt} 0.07 \\
    SVG & -0.08 \hspace{13pt} -0.04 & \textbf{-0.12} \hspace{13pt} \textbf{-0.13} & -0.11 \hspace{13pt} -0.08  \\
    ECOND-HM & -0.03 \hspace{16pt} 0.07 & \textbf{-0.38} \hspace{13pt} \textbf{-0.30} & -0.31 \hspace{13pt} -0.11  \\
    Gromov-Hausdorff dist. & -0.13 \hspace{13pt} -0.07 & \textbf{-0.20} \hspace{13pt} \textbf{-0.20} & -0.18 \hspace{13pt} -0.10 \\
    \bottomrule
    \end{tabular}
    \caption{Correlations (Pearson's $r$) between training data quantities and alignment/isomorphism metrics for language pairs.}
    \label{tab:train_corr}
\end{table*}

\subsection{Feature Search and Ablation} \label{feature_search}

\paragraph{Exhaustive Feature Selection}
We look at the optimal set of language-pair-specific features for predicting the five measures of alignment and isomorphism. To do so, we perform exhaustive feature search on linear regression models
with each of the dependent variables being used separately as the regressand. To counter overfitting, we run ten-fold cross-validation\footnote{ A model's fit is simply averaged over the ten cross-validation runs.} and use \textit{adjusted} $r^2$ as the fit criterion, which further penalizes for additional predictors. Adjusted $r^2$ is given by
\begin{align*}
    r_{adj}^{2} = 1 - \frac{(1-r^2)(n-1)}{n-k-1}
\end{align*}

\noindent where $n$ is the sample size (here, $n=5050$) and $k$ is the number of predictors in the model (here, $1 \leq k \leq 13$). In total, we fit $2^{|F|}=2^{13}=8192$ regression models for LaBSE and LASER separately, where $F$ is our feature space.

For interpretability, we aggregate results by tallying the frequency with which each feature appears in a best-feature list\footnote{Note that there are five dependent variables and two models (LaBSE and LASER), so ten total best-feature lists.}—giving model-specific results as well as combined results—which are displayed in Table \ref{tab:feature-selection}. For the combined (LaBSE+LASER) results, \textit{same polysynthesis status} and \textit{combined in-family sentences} are tied as the most popular predictors, with 8/10 best-feature list appearances each. Next in line for combined results is \textit{combined sentences} (6 appearances), followed by a three-way tie between \textit{same word order}, \textit{token-level overlap}, and \textit{geographic distance} (3 appearances). Results are very similar for each model separately, although \textit{same word order} is tied for second place for LASER, alongside \textit{syntactic distance} and \textit{phonological distance} (3 appearances).

These results show that certain (quasi)-linguistic features (in particular, \textit{same polysynthesis status} and \textit{same word order}) are \textit{not} redundant predictors in the presence of training data quantities. Our next analysis examines individual features in terms of the \textit{size} of their marginal contribution to the regression model fit.

\paragraph{Single-step Regression}

To appraise the marginal contribution of each feature to overall regression fit, we perform a single-step ablation experiment where we eliminate features from a full-feature model one at a time. We fit a regression model with all 13 features using ten-fold cross-validation and obtain a baseline $r^{2}_{adj_{\texttt{bsl}}}$. We then compute 
\begin{align*}
    \Delta r_{adj_{\texttt{f}}}^{2} = r^{2}_{adj_{\texttt{bsl}}} - r^{2}_{adj_{\texttt{abl}}}, \\
    r^{2}_{adj_{\texttt{abl}}} \equalhat F \setminus \{f\} \forall f \in F
\end{align*}
\noindent The value of $\Delta r_{adj_{\texttt{f}}}^{2}$ is computed for all features $f$ and with each dependent variable separately as the regressand, for LaBSE and LASER separately. To aggregate results, we look at the \textit{average rank} of each feature according to the ablation experiment, across all five dependent variables.

The top three results for LaBSE and LASER are given in Table \ref{tab:ablation_rankings}. For LaBSE, \textit{same polysynthesis status} and \textit{combined sentences} are tied as the features with the highest predictive contributions (average rank = 2.4), followed by \textit{combined in-family sentences}. For LASER, \textit{combined in-family sentences} tops the list (average rank = 2.4), followed by \textit{same polysynthesis status} and \textit{same word order}. The results of this experiment are similar, but not identical, to those of the previous experiment. They support the same basic conclusion: \textbf{training data is important, but so are agreement in word order and agreement in morphological complexity}, among other features. If training data were a sufficient predictor alone, then removing the aforementioned features from the regression model would either increase the fit or do nothing, which clearly isn't the case.


\begin{table}[t]
    \centering
    \small
    \begin{tabular}{p{3.6cm}p{.7cm}p{.7cm}p{.7cm}}
        & & \textbf{Count} & \\
        \textbf{Feature} & \textbf{LaBSE} & \textbf{LASER} & \textbf{Total}  \\\toprule
        Comb. sentences & \centering \textbf{4} & \centering 2 & \hspace{4pt} \textbf{6} \\ 
        Comb. in-family sentences & \centering \textbf{4} & \centering \textbf{4} & \hspace{4pt} \textbf{8} \\
        Comb. in-subfamily sentences & \centering 1 & \centering 1 & \hspace{4pt} 2 \\
        Same word order & \centering 1 & \centering \textbf{3} & \hspace{4pt} 4 \\
        Same polysynthesis status & \centering \textbf{4} & \centering \textbf{4} & \hspace{4pt} \textbf{8} \\
        Same family & \centering 1 & \centering 2 & \hspace{4pt} 3 \\
        Same subfamily & \centering 1 & \centering 1 & \hspace{4pt} 2 \\
        Token overlap & \centering 2 & \centering 2 & \hspace{4pt} 4 \\
        Character overlap & \centering 0 & \centering 0 & \hspace{4pt} 0 \\
        Geographic distance & \centering 2 & \centering 2 & \hspace{4pt} 4 \\
        Syntactic distance & \centering 0 & \centering \textbf{3} & \hspace{4pt} 3 \\
        Phonological distance & \centering 0 & \centering \textbf{3} & \hspace{4pt} 3 \\
        Inventory distance & \centering 0 & \centering 0 & \hspace{4pt} 0 \\
    \bottomrule
    \end{tabular}
    \caption{The number of times each of the features appeared in the best-feature lists across the five alignment metrics. The top three results (including ties) in each group are in bold.}
    \label{tab:feature-selection}
\end{table}

\begin{table}[t]
    \centering
    \begin{tabular}{llc}
         \multicolumn{2}{c}{\textbf{LaBSE}} \\\hline
         1. Same polysynthesis status  & {2.4} \\
         2. Combined sentences & {2.4} \\
         3. Combined in-family sentences &  {3.6} \\
    \end{tabular}
    \begin{tabular}{llc}
         \multicolumn{2}{c}{\textbf{LASER}} \\\hline
     1. Combined in-family sentences &  {2.4}  \\
         2. Same polysynthesis status & {3.4} \\
        3. Same word order & {3.8}\\
    \end{tabular}    
    \caption{Features with the top three average rankings in the single-step regression ablation experiment. Rankings are based on a feature's marginal predictive contribution relative to other features, and were averaged across all five alignment metrics.}
    \label{tab:ablation_rankings}
\end{table}

\subsection{Controlling for Training Data}
While the previous experiments center around \textit{prediction} of the dependent variables, we bolster our analysis with classical statistical methods that aim to explicitly control for covariates.\footnote{We consult \citet{dror-etal-2018-hitchhikers, achen-2005-garbage} for guidelines on statistical tests (in NLP).} Since we're dealing with categorical features, we use ANCOVA (\textsc{AN}alysis of \textsc{COVA}riance). 


\paragraph{ANCOVA}
We run ANCOVAs separately for LaBSE and LASER and for each of the five dependent variables. We examine the language-pair-specific features, and look at \textit{same word order} and \textit{same polysynthesis status} separately as our ``between'' variables, and \textit{combined sentences, combined in-family sentences}, and \textit{combined in-subfamily sentences} as our three covariates. 
Overall, \textit{same word order} had a statistically significant ($p<<0.05$) effect for 8/10 ANCOVAs, though effect sizes ($\eta_{p}^{2}$) were generally small\footnote{The rules of thumb we use are: $\eta_{p}^{2}=0.01$ (small); $0.06$ (medium); $0.14$ (large).}\cite{cohen-1988-statistical}. \textit{Same polysynthesis status} had a statistically significant effect for 10/10 ANCOVAs, with effect sizes being definitively small except for F1-score and ECOND-HM/average margin score ($\eta_{p}^{2}\approx0.1$-$0.16$ for LaBSE, $\eta_{p}^{2}\approx0.05$ for LASER). These results suggest that although \textit{same word order} and \textit{same polysynthesis status} are some of the more important features, the determinants of cross-linguality in shared embedding space are multifactorial and most features have a relatively small effect when considered individually.

\section{Zero-shot Cases}

The linguistic diversity of the New Testament Bible corpus, combined with the imperfect overlap between the languages in the corpus and those on which LaBSE and LASER were trained, implies a large number of zero-shot cases for our analysis. We can break these cases into two sub-cases. First, there are languages in the Bible corpus without language-specific training data (35 languages for LaBSE, 45 languages for LASER)\footnote{For both LaBSE and LASER, these languages in fact lack \textit{in-family} training data as well, making the effect of resource scarcity even more pronounced.}. But it follows that there are language \textit{pairs} \textsc{xx-yy} for which no training data is present for either \textsc{xx} or \textsc{yy} (595 pairs for LaBSE, 990 pairs for LASER), which we dub the ``double zero-shot'' case. 

\begin{figure}[h]
    \centering
    \includegraphics[scale=0.6]{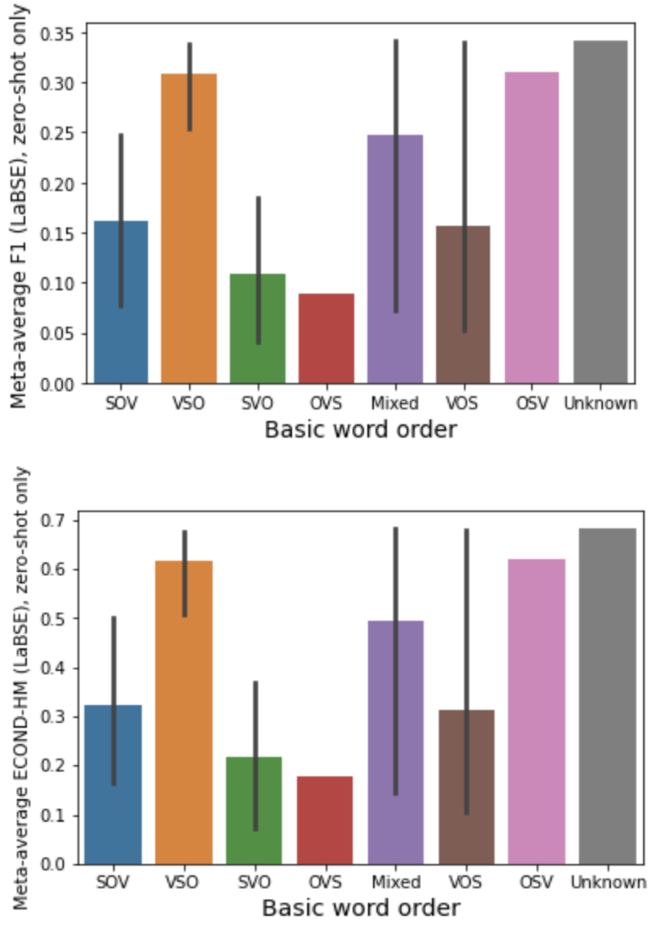}
    \caption{``Meta-average'' performance of zero-shot languages with different word orders on F1-score and negative ECOND-HM (LaBSE).}
    \label{fig:word_order_diffs}
\end{figure}

\paragraph{Simple Zero-shot Case}
For the simple zero-shot case, we use ANOVA (\textsc{AN}alysis \textsc{O}f \textsc{VA}riance) to investigate differences between group means within categorical variables. ANOVAs revealed large effects ($\eta_{p}^{2}\approx0.36$) of \textit{basic word order} on F1-score and ECOND-HM for LaBSE, with borderline p-values ($p\approx0.07$), perhaps due to the small sample size (35 languages). The breakdown across word orders for zero-shot languages is given in Figure \ref{fig:word_order_diffs}. A pairwise Tukey post-hoc test \cite{salkind-2017-post} revealed a borderline-significant difference between SVO and VSO languages, surprisingly in favor of VSO. There were no statistically significant effects of \textit{polysynthesis} for LaBSE or LASER. Interestingly, this suggests that \textit{agreement} in morphological complexity may be important for cross-linguality, but morphological complexity \textit{itself} is not an important factor. More work is needed to validate this conclusion.

ANOVAs also showed large ($\eta_{p}^{2}\approx1$) effect sizes for \textit{family} and \textit{subfamily} membership, though most results were not statistically significant (again, perhaps due to sample size). This suggests that phylogenetic membership still shapes cross-linguality even when training data is perfectly controlled for, which is an interesting finding.

\paragraph{Double Zero-shot Case}

Interestingly, the language-pair-specific feature which stood out most in the double zero-shot case was \textit{inventory distance}, an anomaly in our analyses. \textit{Inventory distance} was correlated with $r\approx0.2$-$0.4$ for 4/5 dependent variables for LaBSE and with $r\approx0.13$-$0.14$ for 2/5 dependent variables for LASER. 

However, as \textit{inventory distance} quantifies phonological distance between languages, it could be confounded with surface-level information. To test this hypothesis, we regress it with \textit{character-level overlap} and \textit{token-level overlap} separately. For LaBSE, effects of \textit{inventory distance} remain significant ($p<0.05$) for all dependent variables when regressing with \textit{token-level overlap}, and 4/5 variables when regressing with \textit{character-level overlap}. We wish to verify the importance of this feature in future studies.

\section{Case Study 1: Morphological Segmentation of Inuktitut}

Based on the above results, we conclude that whether a language has the same \textit{polysynthesis status} as another language will affect their success on a cross-lingual task. However, our observations pertain to correlation, not causality. To test this observation further, we run an experiment in which we introduce a causal intervention. If indeed polysynthesis status matters, then we hypothesize that making a language ``less polysynthetic'' will improve alignment with a more analytic language like English. 

To test this hypothesis, we examine the effect of morphological segmentation of Inuktitut on the bitext retrieval task. Inuktitut is a polysynthetic, indigenous language and is completely zero-shot for both our models, in that not even in-family data is provided during pretraining. The intuition behind our experiment is that by morphologically segmenting a polysynthetic language, the ``polysynthesis status'' of the segmented Inuktitut is made closer to that of a more analytic language. If our previous findings are correct, we expect Inuktitut to align better with English post-segmentation.

We use the first 10,000 sentences from the Nunavut Hansard Inuktitut-English parallel corpus \cite{joanis-etal-2020-nunavut} as our bitext. For the Inuktitut half of the corpus, we use both the ``raw'' version and a version that has been pre-segmented with the \texttt{Uqailaut} morphological analyzer\footnote{\url{http://www.inuktitutcomputing.ca/Uqailaut/}}. 

\begin{figure}[t]
    \centering
    \includegraphics[scale=0.6]{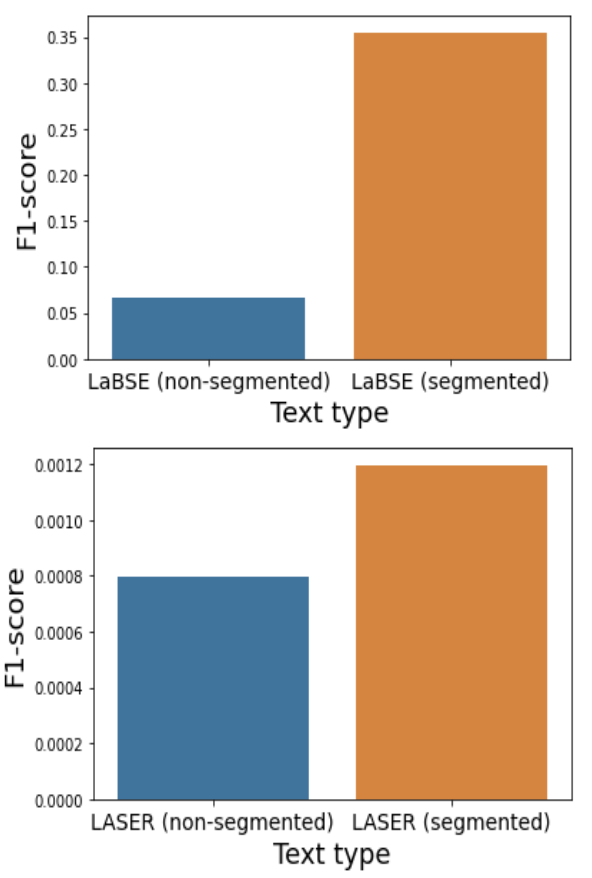}
    \caption{F1-scores on the bitext retrieval task for English-Inuktitut, using raw and morphologically segmented Inuktitut, for LaBSE (top) and LASER (bottom).}
    \label{fig:inuk_f1}
\end{figure}

We then perform bitext retrieval as described in section \ref{bitext-retrieval} on both bitexts: English aligned with non-segmented Inuktitut and English aligned with segmented Inuktitut. Results in terms of F1-score are displayed in Figure \ref{fig:inuk_f1}. For LaBSE, we see a \small$+$\normalsize$28.7$ ($\approx5\times$) F1-score increase using segmented Inuktitut; for LASER, we see a \small$+$\normalsize$0.04$ ($1.5\times$) increase. These empirical results support our earlier statistical findings on the feature \textit{same polysynthesis status}.

\section{Case Study 2: Word Order \& Isomorphism}\label{case_study_2}

To test the validity of our findings on the \textit{same word order} feature, we examine whether embeddings in languages with similar word orders are more isomorphic to each other than those with substantially different word orders, sampling from a different corpus than the one we use in our main analysis. To this end, we select twelve zero-shot\footnote{All languages have no training data for LaBSE or LASER, and most have no \textit{in-family} data either.} languages from the Universal Declaration of Human Rights (UDHR) parallel corpus \cite{vatanen-etal-2010-language}. Six of these are canonically verb-initial: K'iche', Mam, Chinanteco, Tzotzil, Mixteco, and Garifuna. The other six are subject-initial: Chickasaw, Quechua, Achuar-Shiwiar, Bambara, Dagaare, and Guarani. We hypothesize that similar-word-order language pairs will be more isomorphic, on average, than pairs of languages with disparate word orders.

We compute SVG and ECOND-HM across all ${12 \choose 2}=66$ language pairs for LaBSE and LASER separately and group the results based on whether the language pairs have \textit{similar} word order or \textit{different} word order. The averages of these groups are given in Table \ref{tab:udhr_12}. Similar-word-order pairs are more isomorphic than their different-word-order counterparts across \textbf{all} metrics and \textbf{both} models. 




\begin{table}[t]
    \begin{tabular}{lcc}
    & \textbf{SVG} & \textbf{ECOND-HM} \\\toprule
    & LaBSE \hspace{3pt} LASER & LaBSE \hspace{3pt} LASER \\\hline
    \textbf{Similar} & \textbf{5.63} \hspace{3pt} \textbf{3.67} & \textbf{18.08} \hspace{3pt} \textbf{18.13} \\
    \textbf{Different} & 6.34 \hspace{3pt} 4.37 & 18.13 \hspace{3pt} 18.20 \\
    \bottomrule
    \end{tabular}
    \caption{Average values of SVG and ECOND-HM across 66 double zero-shot language pairs in the UDHR subset with \textit{similar} or \textit{different} word orders (based on whether a language is verb-initial or subject-initial). Note that LaBSE and LASER results are not comparable in absolute terms.}
    \label{tab:udhr_12}
\end{table}


\section{Conclusions}

We find evidence that linguistic and quasi-linguistic factors continue to play a role in determining the cross-linguality of a model even after training data is accounted for, and validate our findings with two case studies on extremely low-resource languages. Our analysis points to, among other things, the importance of word order agreement (similarly to \citealt{pires-etal-2019-multilingual}) and morphology agreement on building aligned and isomorphic cross-lingual subspaces. We also rigorously demonstrate the importance of in-family training data in building massively multilingual models, and show moderate effects of other typological measures on cross-linguality. In the future, we are confident that these insights can be used to improve the cross-linguality of shared embedding spaces, particularly for low-resource languages.

\section{Acknowledgements}

We would like to thank Michael Saxon at the University of California, Santa Barbara for his suggestions regarding methodology; Ivan Vulić at the University of Cambridge for his advice on structuring our analysis; and Fangxiaoyu Feng at Google for providing training data statistics on LaBSE.

This work is partly sponsored by the Office of the Director of National Intelligence/Intelligence Advanced Research Projects Activity (IARPA). The views and conclusions contained in this document are those of the authors and should not be interpreted as representing the official policies, either expressed or implied, of the U.S. Government. The U.S. Government is authorized to reproduce and distribute reprints for Government purposes notwithstanding any copyright notation herein.

\section{Ethical Considerations}


When drawing inferences about multilingual language models, it is crucial to take into account languages that are low-resource, Indigenous, and endangered. Previous works have looked at the challenges facing these sorts of under-resourced and under-studied languages (e.g. \citealt{mager-etal-2018-challenges, joshi-etal-2020-state}) and proposed broad solutions and guidelines (e.g. \citealt{kann-etal-2019-towards, bender2019rule}). 


The Bible corpus \citep{christodouloupoulos_steedman_2014} that we use in our analysis includes 35 languages that are zero-shot for LaBSE and 45 that are zero-shot for LASER, all of which could be classified as low-resource or  extremely low-resource. This means that, for our case studies, we can test our conclusions on extremely low-resource languages (including Indigenous languages) that are typically underrepresented in NLP. 

While the Bible corpus enables us to extend our work to low-resource languages, we also acknowledge that the corpus owes its existence largely to a settler colonial tradition, in which missionaries translated the Bible into Indigenous languages---often without crediting the Indigenous peoples who contributed their knowledge. We acknowledge these Indigenous peoples' contributions to this work.


Studies such as \citet{strubell-etal-2019-energy} and \citet{schwartz-etal-2019-green} have identified, analyzed, and proposed solutions for the energy consumption, cost, and environmental impact of NLP models, in particular the burdens associated with training and performing inference with large pretrained language models. Though we perform inference with two such models on a considerable amount of input, we note that these are one-time computations, made using a single NVIDIA V100 GPU, and that we plan to release our collected data publicly for reuse in future empirical analyses.



\bibliography{anthology,custom}
\bibliographystyle{acl_natbib}

\appendix

\section{Appendix}

\subsection{Issues With Using The Bible as a Corpus}

We take note of several issues with using the Bible to perform cross-lingual analyses, but defend our decision to use it over other available corpora. The primary concern is with the language itself of the Bible and its translations: Much of it is archaic and would sound unnatural to modern speakers, and certain translations may suffer from sub-optimal (possibly non-native) translation quality. Furthermore, the relative performance of LaBSE and LASER on these texts was somewhat unrepresentative: LaBSE vastly outperformed LASER, despite the fact that they are closer in performance on more modern, idiomatic texts (e.g. the Tatoeba dataset\footnote{\url{https://github.com/facebookresearch/LASER/tree/master/data/tatoeba/v1}} from \citet{artetxe-schwenk-2019-massively}).

However, the Bible corpus from \citet{christodouloupoulos_steedman_2014} lends itself to our analysis in the following ways:
\begin{itemize}[nosep]
    \item Reliable sentence-level (technically verse-level) alignments
    \item Clean, easy-to-parse text
    \item Large-scale multilinguality and linguistic diversity
\end{itemize}

\noindent We also consider using JW300 \cite{agic-vulic-2019-jw300}, the Tatoeba Challenge test data\footnote{\url{https://github.com/Helsinki-NLP/Tatoeba-Challenge/tree/master/data}} \citet{tiedemann-2020-tatoeba}, and the Johns Hopkins University Bible corpus \cite{mccarthy-etal-2020-johns}. However: 
\begin{itemize}[nosep]
    \item JW300 is difficult to download in its entirety and sentence-align into a superparallel corpus in practice, and alignments may not be as clean as in the Bible corpus
    \item The Tatoeba Challenge bitexts are not multiparallel, so are useless for our main analysis
    \item The Johns Hopkins Bible corpus, while impressive in size with 1600+ languages, is overkill for the intended scale of our analysis (and, in practice, the quality of a corpus of this size is difficult to ascertain)
\end{itemize}
For these reasons, we viewed using the corpus from \citet{christodouloupoulos_steedman_2014} as a ``necessary evil'' of sorts to achieve the scale of analysis we were hoping for.

\subsection{Choice of Embedding Models}\label{sec: choice_of_models}

We opt to use LaBSE \citep{feng-etal-2020-labse} and LASER \citep{artetxe-schwenk-2019-massively} as our embedding models primarily because they are state-of-the-art sentence encoders that perform well on the bitext mining task \citep{reimers-gurevych-2020-making}. Using two models with different underlying architectures (Transformer for LaBSE vs BiLSTM for LASER) makes our analysis more robust and generalizable, because any trend observed w.r.t. both models cannot be due to a peculiarity of one model or the other (e.g. training data domain, neural architecture, tokenization technique, etc.). 

However, while both models have generally high performance on this task, LaBSE is, on average, superior to LASER (see \citet{reimers-gurevych-2020-making}, but also our full results\footnote{\url{https://github.com/AlexJonesNLP/XLAnalysis5K/tree/main/Bible\%20experimental\%20vars}} from this paper). On the lowest-resource languages and language pairs, we see an induced floor effect for LASER, where the variance among data points is low and statistical effects are hard to detect. For the same reason, we do not include results from mean-pooled subword embeddings—such as mBERT or XLM-RoBERTa—due to their relatively weak performance on the bitext mining task \citep{reimers-gurevych-2020-making}.

Floor effects do not pose nearly as much of a problem for LaBSE. Thus, by including LaBSE as one of our models, we are able to detect fine-grained differences among low-resource languages and language pairs that we might miss with LASER. For higher-resource cases, our conclusions are made all the more robust for having inferences from two high-performing models.

\subsection{Principal Component Analysis}

We also perform principal component analysis (PCA) to determine how many independent components exist in our feature space, and how the loadings of those components break down.

\subsubsection{Principal Component Regression}

We run principal component regression (PCR) to determine the optimal number of components in our feature space for predicting the dependent variables. To this end, we first perform PCA on the full set of 13 features (separately for LaBSE and LASER, as the training features are different for each). We then perform PCR (with linear regression) using the first 1 to 13 components in separate runs, with each of the dependent variables being modeled separately as the regressand. As we did before, we measure regression fit using adjusted $r^2$ and average the results from ten-fold cross validation on each run.

We find that for LaBSE, the optimal number of components for predicting the dependent variables averaged 7.2, or roughly half the size of the feature space. For LASER, the average number of optimal components was 6.0.

\subsubsection{Component Loadings}

We also look at how the loadings of the principal components for LaBSE and LASER features break down; the results for the first five components are given in Table \ref{tab:loadings}. For both LaBSE and LASER, the first three components map almost entirely onto training features, while later components are a mixture of the remaining features. However, \textit{same word order} and \textit{same polysynthesis status} are next after training-related features in terms of weight: they are the top two features in components 4 and 5 for both systems.

\subsection{Semi-partial Correlations for Typological Distance}
For the typological distance features, we use the semi-partial correlation \cite{abdi-2007-part}
\begin{align*}
    r_{1(2.3)} = \frac{r_{12}-r_{13}r_{23}}{\sqrt{1-r_{23}^{2}}}
\end{align*}
\noindent where $r_{1(2.3)}$ is the correlation between $f_1$ and $f_2$ such that $f_3$ is held constant for $f_2$ (in our case, training data features are held constant for the dependent variables). This informs us how the typological distance features correlate with the dependent variables when training data features are modeled as covariates. We compute semi-partial correlations between each typological distance measure and each dependent variable for LaBSE and LASER separately.

The typological distance features had noteworthy ($r>0.1$) correlations for anywhere from 0/10 (phonological distance) to 5/10 (geographic distance) analyses. However, the $r$ values generally fell into the range $0.1 < |r| < 0.25$. We conclude that \texttt{lang2vec} distances correlate with cross-linguality weakly but non-negligibly when training data is held constant, somewhat contrary to the stronger relationships observed in \citet{dubossarsky-etal-2020-secret} with monolingual embedding spaces.

\subsection{Visualization from Case Study 2}
\begin{figure*}[h!]
    \centering
    \includegraphics[scale=0.5]{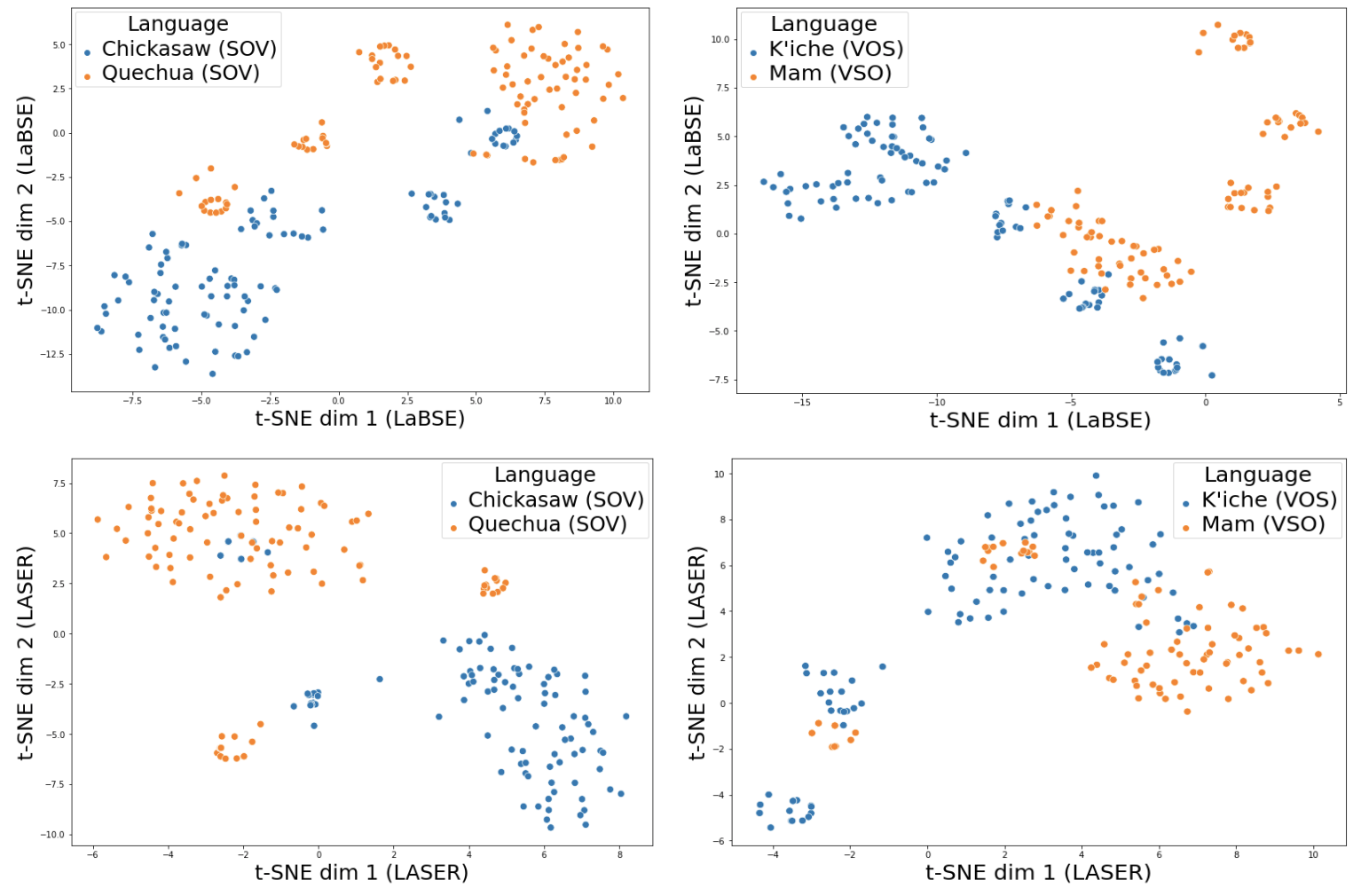}
    \caption{The first two t-SNE dimensions of sentence embeddings in the Universal Declaration of Human Rights, in four zero-shot languages (Chickasaw, Quechua, K'iche, and Mam). Languages with similar word order have been plotted together to demonstrate isomorphism of the resulting vector subspaces (LaBSE plots are top, LASER plots are bottom).}
    \label{fig:word_order_tsne}
\end{figure*}

We visualize approximate isomorphism between select similar-word-order language pairs from section \ref{case_study_2} with t-SNE \cite{maaten-hinton-2008-visualizing}, with default settings in \texttt{scikit-learn}. Results are displayed in Figure \ref{fig:word_order_tsne}.

\subsection{ECOND-HM Computation}
The \textit{condition number} of a matrix $\mathcal{X}$ with $n$ singular values $\sigma_{1}, \sigma_{2}, ..., \sigma_{n}$, sorted in descending order, is defined as:
\begin{align*}
\kappa(\mathcal{X}) = \frac{\sigma_1}{\sigma_n}
\end{align*}
Furthermore, the \textit{effective rank} of $\mathcal{X}$ is defined as:
\begin{align*}
\mathrm{rank}^{*} = \floor{e^{H(\Sigma)}}
\end{align*}
where $\floor{\cdot}$ is the floor function and $H(\Sigma)$ is the \textit{entropy} of the normalized singular value distribution of $\mathcal{X}$, namely $H(\Sigma) = -\sum_{i=1}^{n}{\bar{\sigma_i} \log{\bar{\sigma_i}}}$, where $\bar{\sigma_i} = \frac{\sigma_i}{\sum_{j=1}^{n}{\sigma_{j}}}$. Putting the two together, we define the \textit{effective condition number} of $\mathcal{X}$ as:
$$\kappa_{eff} = \frac{\sigma_1}{\sigma_{\mathrm{rank}^{*}(\mathcal{X})}}$$
Finally, we define the \textit{effective condition number harmonic mean} \cite{dubossarsky-etal-2020-secret} as: 
\begin{align*}
\mathrm{ECOND\_HM}(\mathcal{X},\mathcal{Y}) = \frac{2 \cdot \kappa_{eff}(\mathcal{X}) \cdot \kappa_{eff}({\mathcal{Y})}}{\kappa_{eff}(\mathcal{X})+\kappa_{eff}(\mathcal{Y})}
\end{align*}
\noindent Using the \textit{effective} rank instead of the standard rank to determine the (effective) condition number is a heuristic method motivated by finding the least singular value that characterizes $\mathcal{X}$ in a significant way, as informed by the entropy associated with the singular value distribution of $\mathcal{X}$.

\onecolumn

\begin{table}[h]
    \resizebox{1\textwidth}{!}{
    \centering
    \begin{tabular}{lccccc}
        \textbf{Feature} & \multicolumn{5}{c}{\textbf{Loadings}}\\\hline
          & \textbf{PC1} & \textbf{PC2} & \textbf{PC3} & \textbf{PC4} & \textbf{PC5}  \\
         Combined sentences & \textcolor{red}{2.50e-2}, \textcolor{ForestGreen}{5.28e-2} & \textcolor{red}{-2.30e-1}, \textcolor{ForestGreen}{2.53e-1} & \textcolor{red}{9.73e-1}, \textcolor{ForestGreen}{9.66e-1} & -8.59e-11, -7.83e-10 & 1.12e-11, -5.93e-10 \\
         Combined in-family sentences & \textcolor{red}{9.78e-1}, \textcolor{ForestGreen}{9.60e-1} & \textcolor{red}{2.07e-1}, \textcolor{ForestGreen}{-2.81e-1} & \textcolor{red}{2.38e-2}, \textcolor{ForestGreen}{2.13e-2} & -2.11e-11, -8.33-10 & 2.83e-12, -5.56e-12 \\
         Combined in-subfamily sentences & \textcolor{red}{2.07e-1}, \textcolor{ForestGreen}{2.77e-1} & \textcolor{red}{-9.51e-1}, \textcolor{ForestGreen}{9.26e-1} & \textcolor{red}{-2.30e-1}, \textcolor{ForestGreen}{-2.58e-1} & -5.92e-12, 4.33e-10 & 1.15e-11, 4.40e-10 \\
         Same word order & 1.10e-11, 3.42e-10 & -5.38e-12, -4.97e-10 & 7.59e-11, 1.15e-9 & \textcolor{red}{7.60e-1}, \textcolor{ForestGreen}{7.13e-1} & \textcolor{red}{6.17e-1}, \textcolor{ForestGreen}{6.73e-1} \\
         Same polysynthesis status & 1.57e-11, 4.30e-10 & -2.21e-11, -8.02e-11 & 6.69e-11, 1.11e-10 & \textcolor{red}{6.02e-1}, \textcolor{ForestGreen}{6.54e-1} & \textcolor{red}{-7.81e-1}, \textcolor{ForestGreen}{-7.33e-1} \\
         Same family & 2.44e-11, 6.97e-10 & -1.19e-12, -1.28e-10 & -2.56e-11, -4.76e-11 & \textcolor{red}{1.67e-1}, \textcolor{ForestGreen}{1.84e-1} & -6.07e-4, -1.11e-2 \\
         Same subfamily  & 3.64e-12, 1.10e-10 & -1.48e-11, 1.19e-10 & -1.81e-11, -7.06e-11 & 7.38e-2, 7.71e-2 & 1.45e-2, 1.65e-2 \\
         Token overlap & 6.68e-12, 2.07e-10 & -1.42e-11, 2.67e-11 & -1.20e-11, 6.42e-12 & 6.23e-2, 6.48e-2 & -1.87e-2, -1.77e-2 \\
         Character overlap & 2.53e-12, 1.72e-10 & -2.34e-11, 3.68e-10 & -1.02e-10, -8.57e-11 & 4.93e-2, 2.39e-2 & 2.71e-2, 3.63e-2 \\
         Geographic distance & -4.28e-12, -1.29e-10 & 6.74e-13, 4.58e-11 & -7.91e-13, -5.54e-11 & -6.41e-2, -6.77e-2 & \textcolor{red}{6.45e-2}, 6.20e-2 \\
         Syntactic distance & -7.48e-12, -2.36e-10 & 6.14e-12, 1.02e-10 & 1.53e-11, -2.23e-11 & -9.85e-2, -9.33e-2 & -6.25e-2, \textcolor{ForestGreen}{-6.48e-2} \\
         Phonological distance & -5.80e-12, -1.70e-10 & -3.28e-12, 1.49e-10 & 2.87e-11, -9.04e-12 & -5.81e-2, -5.71e-2 & 2.20e-2, 2.19e-2 \\
         Inventory distance & -8.39e-13, -2.12e-11 & -3.53e-12, 1.05e-10 & 1.64e-11, -3.90e-11 & -4.75e-2, -4.51e-2 & 1.49e-2 1.23e-2 \\
    \bottomrule
    \end{tabular}}
    \caption{Loadings from the first five principal components for the language-pair-related features. The top three loadings by magnitude in each component are colored \textcolor{red}{red} for LaBSE and \textcolor{ForestGreen}{green} for LASER. Note that although LaBSE and LASER are trained using different neural architectures, the most significant features in each of the first five components are almost identical.}
    \label{tab:loadings}
\end{table}

\end{document}